\documentclass[11pt]{article}
\usepackage{coling} 
\usepackage{times} 
\usepackage[utf8]{inputenc}
\usepackage[T1]{fontenc}
\usepackage{graphicx}
\usepackage{grffile}
\usepackage{longtable}
\usepackage{wrapfig}
\usepackage{rotating}
\usepackage[normalem]{ulem}
\usepackage{amsmath}
\usepackage{textcomp}
\usepackage{amssymb}
\usepackage{capt-of}
\usepackage{hyperref}

\usepackage{todonotes}
\usepackage{covington}
\usepackage{color,soul}
\usepackage{comment} 

\usepackage{todonotes}

\usepackage{enumitem}
\usepackage{array} 
\usepackage{pgfplots} 
\usepackage{multirow} 
\title{Controlling Out-of-Domain Gaps in LLMs for Genre Classification and Generated Text Detection}

\hypersetup{
 pdfauthor={},
 pdftitle={Controlling Out-of-Domain Gaps in LLMs for Genre Classification and Generated Text Detection},
 pdfkeywords={},
 pdfsubject={},
 pdfcreator={Emacs 27.2 (Org mode 9.4.4)}, 
 pdflang={English}}

\author{Dmitri Roussinov\\
  University of Strathclyde\\16 Richmond Street, Glasgow G1 1XQ\\
  \texttt{dmitri.roussinov@strath.ac.uk}
  \And
  Serge Sharoff \\
  University of Leeds\\ Leeds LS2 9JT\\
  \texttt{s.sharoff@leeds.ac.uk}
  \AND
  Nadezhda Puchnina \\
  Independent consultant\\
  \texttt{nadezhdapuchnina35@gmail.com}
}

\begin{document}

\maketitle
\begin{abstract}

This study demonstrates that the modern generation of Large Language Models (LLMs, such as GPT-4) suffers from the same out-of-domain (OOD) performance gap observed in prior research on pre-trained Language Models (PLMs, such as BERT). We demonstrate this across two non-topical classification tasks: 1) genre classification and 2) generated text detection. 
Our results show that when demonstration examples for In-Context Learning (ICL) come from one domain (e.g., \textit{travel}) and the system is tested on another domain (e.g., \textit{history}), classification performance declines significantly.

To address this, we introduce a method that controls which predictive indicators are used and which are excluded during classification. For the two tasks studied here, this ensures that topical features are omitted, while the model is guided to focus on stylistic rather than content-based attributes. This approach reduces the OOD gap by up to 20 percentage points in a few-shot setup. Straightforward Chain-of-Thought (CoT) methods, used as the baseline, prove insufficient, while our approach consistently enhances domain transfer performance.


\end{abstract}

\section{Introduction}
\label{sec:org2ffd441}

Recent advancements in Large Language Models (LLMs) have pushed the boundaries of natural language processing, leading to remarkable performance across a wide range of tasks \cite{brown2020language, floridi2020gpt, bubeck2023sparks}. While their success with In-Context Learning (ICL) has gained particular attention, questions remain regarding their consistency when applied to unfamiliar domains. Unlike models similar in size to BERT \cite{devlin2018bert}, which require a sample of labeled data for domain-specific tuning, LLMs are often used via ICL with little to no fine-tuning. However, this flexibility comes at a cost, as LLMs frequently experience a decline in performance when tested across domains \cite {yuan2024revisiting}, a gap also observed in earlier, smaller models \cite{roussinov23emnlp}. This degradation is often attributed to the models' reliance on surface-level features rather than deeper, domain-independent attributes \cite{wang23icl}. 

To address this challenge, we introduce a method that controls which prediction indicators the model considers during few-shot document classification. While this approach is broadly applicable to many non-topical document classification tasks, our focus here is on the ICL approach to two specific tasks: 1) the automated recognition of document genre, and 2) detection of computer-generated texts.  

Genre classification plays a vital role in fields such as information retrieval, automatic summarization \cite{stewart2009genre}, machine translation \cite{van2018evaluation}, and dependency parsing \cite{muller2021genre}. It also aids information security by enabling genre-aware assessments of web document credibility \cite{agrawal2019fact}, and is crucial for curating genre-diverse corpora to pre-train LLMs to build robust foundation models \cite{kuzman2023automaticLLM, lepekhin2022estimating}. While recent advancements in LLMs have shown that zero-shot methods can achieve strong performance in genre classification \cite{kuzman2023automaticLLM}, another qualitative exploratory study pointed out that a few-shot approach may severely suffer from OOD performance gaps \cite{roussinov23emnlp}. 

Detecting AI-generated text has become critical in verifying authenticity and maintaining information integrity. As models like GPT-4 produce increasingly human-like text, the risks of misinformation, plagiarism, and malicious content generation have surged. The task involves distinguishing human-authored text from machine-generated outputs, leveraging techniques that identify linguistic patterns or use methods like watermarking (embedding hidden markers in generated text). Recent surveys emphasize the importance of detecting generated content for preventing deepfakes, maintaining trust in digital communications, and ensuring transparency in AI applications \cite{tang2024science, gehrmann2019gltr, kirchenbauer2023watermark}. 

Here are our specific contributions:\footnote{ 
{For replicability of our experiments, the full experimental setup is at} https://github.com/dminus1/LLM-OOD-control}
\begin{enumerate}
\item While prior studies focused on datasets from distinct sources or smaller models like BERT \cite{kuzman2023automaticLLM, roussinov23emnlp}, our work directly \textit{confirms significant out-of-domain performance gaps} in LLMs across two key tasks: genre classification and generated text detection, both using In-Context Learning across multiple topics (domains). We evaluated these tasks using two advanced LLM families (GPT-4\footnote{https://chat.openai.com/} and Claude\footnote{https://claude.ai/}), accessed through their APIs.

\item Earlier studies with smaller (BERT-sized) pre-trained language models (PLMs) reported only modest reductions in OOD gaps (e.g., 2?3 percentage points; \citealp{roussinov23emnlp}). In contrast, our results demonstrate significant improvements, with reductions of up to 7 and 20 percentage points across the two tasks, respectively.
This facilitates domain transfer by enabling classifiers demonstrated with few-shot examples from one domain (e.g., \textit{travel}) to be effectively applied in another (e.g., \textit{history}). Our method is distinctive in its ability to control which text attributes are emphasized (e.g., style) and which are disregarded (e.g., topical content). 

\item Through ablation studies, we verify that detailed prompts are crucial for optimal performance, while the straightforward application of Chain-of-Thought (CoT, \citeauthor{wei2022chain}, \citealp{wei2022chain}) lacks the precision needed to guide the model effectively.
\end{enumerate}



\section{Methodology}  
\label{sec:org51d846c}
\begin{figure*}[!t]
\centering
\includegraphics[width=6.0in]{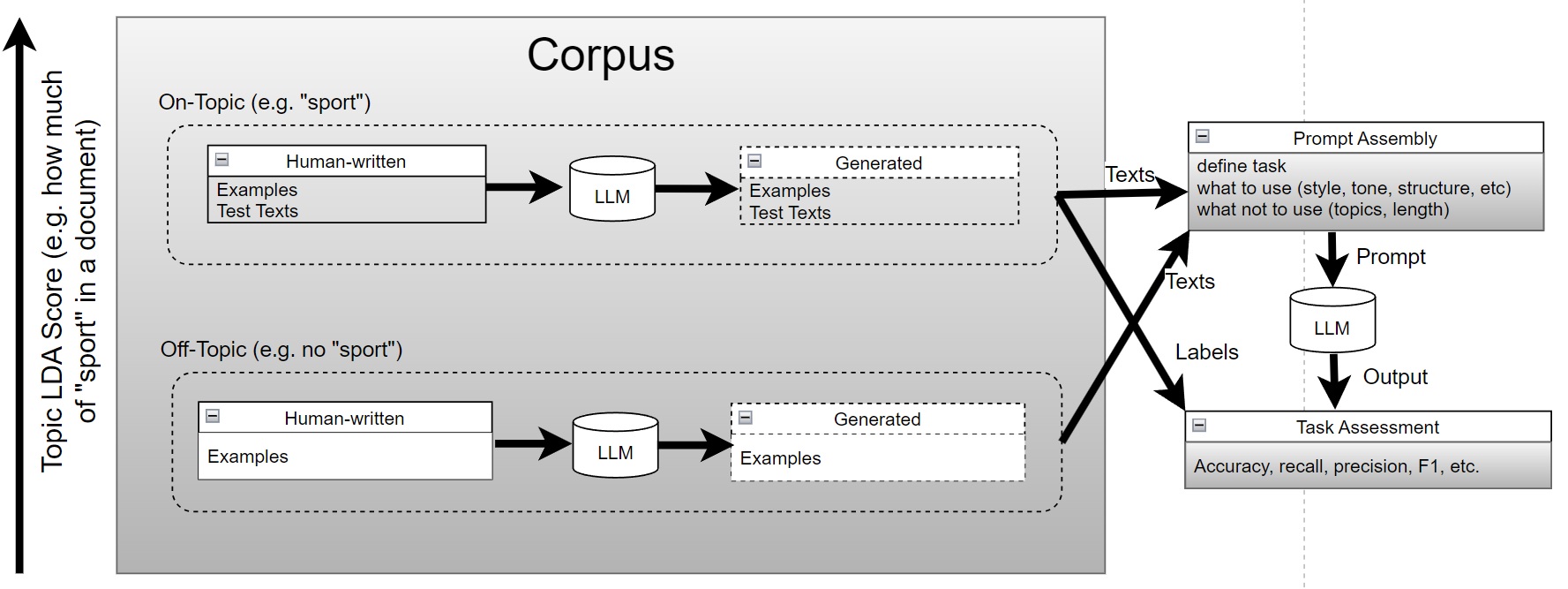}
\caption{Domain transfer assessment methodology adapted from \cite{roussinov23emnlp} for few-shot In-Context Learning (ICL), independently testing two tasks: (1) genre classification and (2) generated text detection. Prompt construction may optionally include instructions on which indicators to use or ignore. The topic modeling scores determine which texts are considered on-topic (top) or off-topic (bottom). On-topic texts are used for testing and, depending on the configuration, for ICL demonstration examples, while off-topic texts are only used as examples. Synthetic texts are generated by an LLM for the generated text detection task. This methodology is applicable to other non-topical classification tasks, such as determining gender, identifying authorship, analyzing sentiment, etc.}
\label{figDiagram}
\end{figure*}

\subsection{The Approach} 

Figure \ref{figDiagram} illustrates the overall workflow for our experiments. As in prior recent works on genre classification, e.g., \newcite{kuzman2023automaticLLM} or \newcite{roussinov23emnlp}, we define genre as the 'function of the text, author?s purpose, and form of the text.' Writing style is an important characteristic of genre but not the only one. Other characteristics include the intended audience, the medium through which the text is delivered, and the context of its use.

Following the methodology in \newcite{roussinov23emnlp}, we define 'domain' as a topic in the topic modeling sense \cite{blei03}. Our approach, therefore, focuses on identifying distinct topics such as \textit{sports, politics, or health} from a large general-purpose corpus. This contrasts with several previous studies (e.g., \citealp{kuzman2023automaticLLM}), where 'domain' is defined by differences in dataset collection or labeling methods. For example, in their approach, book reviews and movie reviews would be considered separate domains within a sentiment analysis task. 

We employ the domain transfer assessment methodology and datasets from \newcite{roussinov23emnlp}, originally developed to test out-of-domain (OOD) classification with BERT-sized PLMs, to further investigate and address the OOD gap in few-shot genre classification using ICL and large language models (LLMs). 
We propose and validate a domain transfer approach by adapting Chain-of-Thought (CoT, \citealp{wei2022chain}) prompting to control which text properties should be emphasized (e.g., writing style, purpose) and which should be ignored (e.g., specific topics). To the best of our knowledge, no prior studies have explored such control in this context.

From our previous study \cite{roussinov23emnlp},  we borrow the topic model estimation, the corpus of documents to classify, and the evaluation mechanism for the OOD effect. The core idea is to simulate a scenario where a classifier is shown documents which are far away from a particular topic (e.g., \textit{sports}) and then its performance is evaluated on documents where that topic is prominent. 
This performance is compared to a scenario where the classifier is shown documents in which the same topic is prominent (\textit{on-topic}). While our findings focus on genre classification and detection of computer-generated texts, this methodology is flexible and can be easily adapted to other non-topical classification tasks, such as determining gender, identifying authorship, or analyzing sentiment. Further details are provided in the following subsections.

\subsection{Corpus}
The corpus from our previous study \cite{roussinov23emnlp} provides good coverage of several genres and topics.
Up to our knowledge, there is no other large corpus for that purpose. The corpus has been collected via ``natural genre annotation'' by combining several sources so that each source is relatively homogeneous with respect to its genres.  The description of the genre classes follows prior studies of genre types common on the Web \cite{sharoff18genres}.  The composition of the natural genre corpus is listed in Table \ref{tabCorporaG} in Appendix. 

Our topic model was trained on ukWac, a much bigger topically diverse corpus \cite{baroni09}, to infer themes across all sources of our natural genre corpus.  In this way, we obtain two complementary perspectives on each document: its topic and its genre. For example, take the following excerpt:

\newenvironment{nonumberexamples}
  {\begin{list}{}{\leftmargin=0pt \itemindent=0pt \itemsep=0pt}}
  {\end{list}}

\begin{nonumberexamples}
\small 
\item Following Mary Smith's thorough review of this album, there?s not much left to add, but I was so moved by the music that I had to contribute my thoughts. Her review highlights key tracks like Miles Davis' "So What" and John Coltrane's "Blue Train" \ldots
\end{nonumberexamples}

\noindent This document is classified as a Review (originating from the Amazon Review collection) and is linked to Topic 1 (\textit{entertainment}, see the labels in \autoref{tabCorporaT}) based on our topic model. Reviews can span various topics, such as \textit{science} or \textit{history}. This dual classification?by both genre and topic?enables us to effectively assess OOD performance across different domains.

\subsection{Domain Transfer} 
To test the effect of a topic change  we also used the methodology suggested by \newcite{roussinov23emnlp}, which is briefly summarized in this subsection. While developed specifically for document genres, this methodology is applicable to any non-topical classification, so it has been tested here on the task of detecting computer-generated texts.  

We make the following distinction between  \textit{on-topic} and \textit{off-topic} examples, e.g. \textit{sport}.
The highest scoring documents,  according to the topic model, are designated as \textit{on-topic} examples for each genre. Additional (non-overlapping with the ICL examples) highest scoring documents are used as test cases (test-set), associated with that particular topic. The lowest scoring documents are designated as \textit{off-topic} examples. In our example, those would be the documents that are definitely not about \textit{sports}. The methodology contrasts the performance metrics between scenarios using {off-topic} or {on-topic}  examples. 

\subsection{Dataset for Generated Text Detection}
Since publicly available datasets for generated text detection do not support testing for out-of-domain (OOD) gaps and transfer, we synthesized our own datasets. Using the same corpus as in our genre classification experiments, we created off-topic and on-topic datasets for generated text detection. During this process, we excluded the PERSonal genre from the corpus due to occasional adult content that triggered API warnings in our preliminary experiments (while the API did not object to analyzing such texts, it refused to generate them). Additionally, we excluded the INFOrmation and INSTRuction genres because the generated texts based on these were structurally distinct from the originals, making it trivial for both humans and models to distinguish between the generated and original texts.

We tasked Claude 3 Sonnet with generating text "on the same topic and in the same style" as the texts from off-topic and on-topic documents in the genre corpus, ensuring a balanced distribution across the three remaining genres (ARGument, NEWS, and Review) to maintain diversity. This process produced two synthesized datasets for each topic: one with on-topic demonstration examples and one with off-topic demonstration examples, both sharing the same on-topic test texts. Each dataset included 5-shot demonstration examples and 10 test cases per topic, mirroring the sizes used in our genre classification task.

\subsection{Metrics}
While \newcite{kuzman2023automaticLLM} assessed LLMs as genre classifiers through a multi-class task, we followed the approach in \newcite{roussinov23emnlp}, who assessed ChatGPT through \textit{binary} classification between pairs of genres. 
This approach reduces the number of examples in our prompts, allowing them to fit within the current context window limits of the models used, and to keep the costs reasonable\footnote{ The total cost of using the LLM APIs was approximately 200 US dollars.}. From the evaluation viewpoint, this formulation is methodologically equivalent to multi-class classification \cite{allwein2000reducing,vapnik2013nature}. For each topic, we randomly select (without replacement) a pair of genres and test the binary classification accuracy using a balanced test set consisting of 10 randomly selected texts (5 of each genre). These texts are randomly sampled from the smallest dataset provided by \newcite{roussinov23emnlp}. Thus, we tested 25 pairs (one for each topic), which exceeds the total number of unique pairs (15) available when selecting from 6 genres.

We report only accuracy as the comparison metric since our test sets are perfectly balanced, making accuracy both a fair and straightforward metric to interpret. Additionally, obtaining the area under the ROC curve would require ranking the predictions, which would further complicate the LLM?s task. Hence, we decided to avoid it after our preliminary investigations. These investigations also helped us determine that using five examples per prompt, referred to as \textit{five-shot}, was the best compromise between the prompt size and performance.

For genre classification, we used the same document content as in \newcite{roussinov23emnlp}, which are randomly positioned windows of 1000 characters. This mitigates the impact of the document structure, e.g. an introductory question positioned at the start of each document in StackExchange. Their reported experiments with human raters show that the windows obtained this way still provide sufficient information to determine the topic and genre with accuracy around 90\%. To construct the dataset for the detection of computer-generated texts, we used the first 1000 characters from each original text. 

\subsection{Prompts}
\label{sec:orgda42efb}

We tested the configurations (prompt types) described in the subsections immediately below to evaluate the classification accuracy with the following models through their application interfaces: GPT-4o, GPT-3.5, Claude 3 Opus, Claude 3 Sonnet, Claude 3.5 Sonnet, and Claude 3.5 Haiku\footnote{Information about the number of parameters can be found at \url{https://platform.openai.com/docs/overview} and \url{https://claude.ai/}.}.  
These configurations are designed to progressively add more control over which indicators the model should prioritize. The prompts were developed based on our preliminary experiments and informed by prior works \cite{crowston10,rehm2008towards, stein2011web}. The same prompts were used consistently across all the models without any modifications. We omitted class descriptions to simulate realistic few-shot learning scenarios where LLMs must infer task structure from examples alone. Including class descriptions would test the LLM's pre-trained knowledge of these classes (e.g. specific genres here) rather than its ability to generalize from provided examples. Our focus was on assessing how well the model transfers across different topics (domains) without relying on explicit associations.

\subsubsection{Baseline Prompt} 
As the baseline prompt for the genre classification task, we used a Chain-of-Thought (CoT) prompt \cite{wei2022chain} 
without specifying which document features to consider for classification. Since our focus is on large language models of GPT-3 size and larger, for which fine-tuning is prohibitively costly, we consider CoT prompting to be a strong and relevant baseline for these models. CoT prompts are particularly practical given the constraints and objectives of our study. CoT prompting was applied as a two-stage process: (1) instructing the LLM to articulate criteria for distinguishing between classes using the examples provided, and (2) asking the LLM to apply these criteria to classify the test texts.
For the generated text detection task, the baseline prompt simply asked the model to classify texts based on the provided examples (single-stage). 


\subsubsection{Prompt with Simple Control}
In genre classification, the prompts explicitly instruct the model to classify based on document genre without defining what "genre" is. For detecting computer-generated texts, the prompts simply add the instruction: "When classifying, don't use the topic of the text as a criterion" to the baseline prompt.
\subsubsection{Prompt with Detailed Control}
In genre classification, the detailed prompts instruct the model to focus on stylistic and structural indicators such as formality, tone, sentence structure, language complexity, purpose (e.g., to inform, instruct, or facilitate dialogue), use of perspectives (first, second, or third person), active voice, and features like citations, references, or personal experiences. The Detailed Control prompts for both tasks explicitly prohibit using topical content or text length as classification criteria, emphasizing that the analysis should remain universally applicable across all topics. For instance, the instructions state: "Your criteria should not mention any specific topics and should be applicable to the texts on ANY topic!" The prompts also list examples of possible topics from our topic model, including "business, finances, entertainment, universities, markets, science, politics," and others.


\section{Results and Discussion} 
\label{sec:org6ad1310}
\begin{table*}[h]
\caption{\textbf{Binary classification accuracy (\%) for various prompt configurations in a 5-shot setting.} The ``Basic Prompt'' (baseline) instructs the model to classify documents based solely on the provided examples, without specifying which indicators to consider. The ``Prompt with Simple Control'' instructs the model to avoid using the topical content of the texts as classification criteria, thus facilitating domain transfer. The ``Prompt with Detailed Control'' (empirically found to be the best) specifies which indicators to prioritize (e.g., style, purpose, structure) and provides explicit examples of topic-based indicators to avoid. }
\label{tabResultsMerged}
\setlength{\tabcolsep}{4pt}
\centering
\small
\begin{tabular}{p{4cm}|c|ccc}
\hline  
\multirow{4}{*}{\textbf{Model \& Task}} & on-topic & & off-topic &  \\
                                        & examples & & examples  &  \\
\cline{2-5}                                        
                                        & basic  & basic prompt  & prompt & prompt \\
                                        & prompt    & (baseline) & with simple control & with detailed control \\
\hline  
\textbf{Genre Classification}: & & & & \\ 
GPT-4o          & 90.4 & 68.8 & 70.0 & \textbf{76.0} \\
Claude 3 Opus     & 85.6 & 72.0 & 77.2 & \textbf{78.4} \\
Claude 3.5 Sonnet & 84.4 & 75.6 & 81.6 & \textbf{82.0} \\ 
Claude 3.5 Haiku  & 79.6 & 70.8 & 69.6 & \textbf{72.0} \\ 
GPT-3.5         & 71.2 & 60.8 & 62.8 & \textbf{63.2} \\
Claude 3 Sonnet   & 75.6 & 64.4 & 65.6 & \textbf{66.0} \\

\hline  
\textbf{Generated Text Detection}: & & & & \\ 
GPT-4o        	&82.8 & 65.2 & 68.4 & \textbf{85.2}    \\
Claude 3.5 Haiku &87.6 &73.2 &76.4 &\textbf{82.0} \\ 
GPT-3.5     	&64.8 & 64.8 & \textbf{67.2} & 66.0    \\
Claude 3 Sonnet 	&77.2 & 68.4  & 69.2 & \textbf{76.4}   \\
\hline  

\end{tabular}
\end{table*}

\begin{table*}[h]
\caption{Classification accuracy for the \textbf{ablated} genre classification and \textbf{re-phrased} versions of our prompts with GPT-4o.
\label{tabResultsAblation}}
\setlength{\tabcolsep}{4pt} 
\centering 
\small
\begin{tabular}{l|ccc|p{1.1cm}p{1.1cm}p{1.1cm}|p{1.1cm}p{1.1cm}p{1.1cm}} 
& \textbf{without }  & \textbf{without} & \textbf{only} & \multicolumn{3}{c}{\textbf{detailed control prompt rephrased}}   & \multicolumn{3}{c}{\textbf{baseline prompt rephrased}} \\
& \textbf{defining}  & \textbf{listing} & \textbf{half of} & & & & &    \\
& \textbf{topical }  & \textbf{genre } & \textbf{genre} & \textbf{} &  &   &&&\\
\textbf{Topics}:& \textbf{features}  & \textbf{features} & \textbf{features} & run 1 & run 2 & run 3& run 1 & run 2 & run 3\\
 \hline  
			&	74.4	&	72.8	&	75.2	&	75.6	&	75.6	&	76.4	&	68.4	&	68.0	&	67.6	\\
&		\\
\end{tabular}
\end{table*}

The results from both tasks are presented in Table \ref{tabResultsMerged} and are discussed in detail in the following subsections.

\subsection{Confirming OOD Gap with LLMs} The "Basic" prompt shows significantly worse performance across all models and tasks when off-topic examples are used compared to on-topic examples (except for GPT-3.5 in the generated text detection task, where both results are consistently low). These differences are statistically significant at the 0.05 level, as confirmed by the McNemar test \cite{dror2018hitchhiker}. While this finding aligns with previous studies \cite{kuzman2023automaticLLM, roussinov23emnlp}, our work is the first to methodically quantify the OOD gap in few-shot ICL prompts. Additionally, we extend these findings by proposing a remedy for this gap, with results discussed in the following subsection. 
\subsection{Positive Impact of Control}
Applying even a basic level of control over which indicators the models should prioritize resulted in classification accuracy improvements across most models, with gains of up to 6\% on both tasks. By guiding the models to focus on relevant features while disregarding misleading topical cues, we observed enhanced performance in both genre classification and generated text detection. These results suggest that even modest interventions in how models interpret input can significantly reduce out-of-domain performance gaps.
\subsection{Importance of Detailed Control}
For the genre classification task, the "Prompt with Detailed Control" achieved the highest accuracies with the more powerful models (GPT-4o, Claude 3 Opus, Claude 3.5 Sonnet), reducing the OOD gap by approximately 33\% (relatively) for GPT-4o and nearly 50\% (relatively) for Claude Opus. In the generated text detection task, detailed control completely eliminated the OOD gaps for two models (GPT-4o and Claude 3 Sonnet) out of the four tested. For Claude 3.5 Haiku, the OOD gap was reduced by 60\% (relatively), while the oldest model (GPT-3.5) showed the lowest performance and exhibited no sensitivity to the OOD gap. These results underscore that \textit{explicitly specifying which indicators to prioritize significantly enhances the models? ability to handle classification tasks, even with off-topic examples.} Additionally, we observed a correlation between reduced reliance on topical criteria in the LLM outputs and improved accuracy, a trend more pronounced in the larger and more recent models.
 
As observed in previous studies, more powerful models (GPT-4o, Claude 3 Opus, Claude 3.5 Sonnet) consistently outperformed less powerful models (GPT-3.5 and Claude 3 Sonnet) across nearly all prompt configurations and tasks, particularly with detailed instructions. While this result is unsurprising, our findings provide specific numerical ranges for these improvements in the OOD context. These findings emphasize that: (a) the OOD problem remains unresolved, as performance often falls short of the levels achieved with on-topic examples, and (b) no saturation in performance gains has been observed with increasing model size. 

\subsection{Ablation Studies} 
The ablations targeted key components of the prompt: (1) removing the definition and examples of "topical" features, (2) omitting the explicit listing and examples of style/genre-related features allowed for classification, and (3) reducing the descriptions of these features by half. The results show a significant performance drop?more than half?towards baseline levels, with the omission of genre-related features having the most pronounced negative impact on accuracy.

To ensure that our findings were not overly dependent on the specific wording of the prompts, we followed the methodology of \cite{kirchenbauer2023reliability} and used GPT-4o to paraphrase our prompts. As seen in the last six columns of Table \ref{tabResultsAblation}, while minor variations in accuracy occurred with the rephrased prompts, the overall comparative trends remained stable, confirming the robustness of our observations.

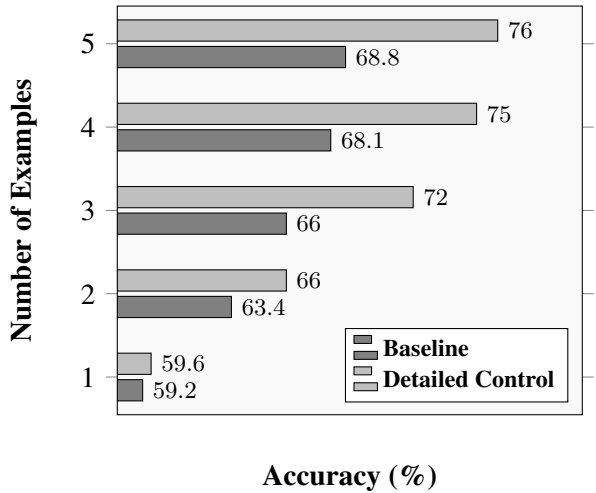
\begin{figure}[htb]
  \centering
  \begin{tikzpicture}
    \begin{axis}[
      xlabel = {\textbf{Accuracy (\%)}}, 
      ylabel = {\textbf{Number of Examples}}, 
      xbar, 
      bar width=8pt, 
      width = \linewidth, 
      height = 7cm, 
      xmin = 58, xmax = 80, 
      ytick = data, 
      enlarge y limits={abs=0.5cm}, 
      xtick=\empty, 
      yticklabels={1, 2, 3, 4, 5}, 
      legend style = {
        at={(0.97,0.03)}, 
        anchor=south east, 
        font=\small 
      },
      legend cell align={left}, 
      axis background/.style={fill=gray!5} 
    ]

      \addplot+[
        xbar, 
        fill=gray, 
        draw=black, 
        nodes near coords, 
        every node near coord/.append style={black, anchor=west, font=\small}, 
      ] coordinates {(59.2,1) (63.4,2) (66,3) (68.1,4) (68.8,5)};
      
      \addplot+[
        xbar, 
        fill=gray!50, 
        draw=black, 
        nodes near coords, 
        every node near coord/.append style={black, anchor=west, font=\small}, 
      ] coordinates {(59.6,1) (66,2) (72,3) (75,4) (76,5)};
      
      \legend{\textbf{Baseline}, \textbf{Detailed Control}} 
      
    \end{axis}
  \end{tikzpicture}
  \caption{Accuracy comparison between GPT-4o baseline and detailed control prompts across different numbers of demonstration examples (shots).}
  \label{fig:plot}
\end{figure}

Figure \ref{fig:plot} illustrates the reduction in the OOD gap for GPT-4o as the number of examples decreases, highlighting that our selection of 5-shot examples strikes a reasonable balance between token cost and the exploration of OOD effects. Although using more examples could enhance performance in practical scenarios, this would also demand significantly greater labeling efforts for each genre and domain, which could extend to hundreds of domains and genres \cite{crowston10}.

We also compared our results by fine-tuning BERT \cite{devlin2018bert} using the same 5-shot setup. To ensure a fair comparison, we did not use validation (development) sets, as that would require more than 5 labeled examples. Instead, we ran training for 
{200} epochs,
starting with a learning rate of 0.00001 and gradually reducing it to 0. The accuracy obtained for genre classification with off-topic examples was 64.4\%, and for generated text detection, 62.0\%. These results are below those of the Large Language Models (GPT-4o and Claude 3 Opus), aligning with previous comparisons reported in \cite{kuzman2023automatic}. We experimented with various learning rate and epoch settings, but results remained consistent.

\section{Related Work}
\label{sec:orgbab98a2}
While it has been noted that zero-shot and few-shot ICL based on LLMs suffers less from the OOD gap in comparison to fine-tuned smaller PLMs, there are no universally successful solutions for domain transfer yet \cite{yuan2024revisiting, edwards24incontext}.

The use of ICL for domain adaptation in general has been explored in \cite{long2023adapt}, which focused on additional pre-training on the target corpus and selecting the most similar examples (demonstrations) from the source domain. Our work differs in several key aspects: 1) We specifically investigate large language models (GPT-3.5 and larger), whereas \newcite{long2023adapt} primarily used smaller models, except for GPT-3.5-turbo, with which they reported negative results. 2) We operate under the constraint of having no more than five examples available ("true" few-shot), eliminating the need for example selection, which in general could serve as an additional avenue for improvement.


\subsection{Automated Genre Classification}

For a recent survey on genre classification, including fine-tuned smaller-size PLMs and ICL use of Large Language Models, we refer the reader to the work by \newcite{kuzman2023automatic}. 
They note that genre classification is an important task as it relates to the very "purpose" of the text. For instance, distinguishing a document genre with a high degree of humor  (e.g. an \textit{anecdote}) from the one with factual information is crucial for its proper interpretation. Consequently, obtaining genre information has been shown to be beneficial for a wide range of disciplines, including \textit{corpus linguistics, computational linguistics, natural language processing, information retrieval,} and \textit{information security}.
Additionally, curating corpora with a variety of genres for pre-training  LLMs themselves is essential to ensure robust and comprehensive foundation models \cite{kuzman2023automaticLLM,lepekhin2022estimating}. 
As noted by \newcite{kuzman2023automaticLLM}, while BERT-sized models demonstrate exceptional performance in genre classification, significantly outperforming earlier SVM and other classical machine learning approaches, they still require a considerable amount of labeled texts for fine-tuning. Recent advancements have shown that instruction-tuned GPT-like generative models \cite{brown2020language}, when used in zero-shot or few-shot settings, can achieve comparable or even superior results without the need for large-scale labeling efforts. Building on the work of \newcite{kuzman2023automaticLLM} and \newcite{roussinov23emnlp}, our focus is on facilitating domain transfer in a few-shot setting by applying In-Context Learning (ICL) with GPT-3.5-sized or larger LLMs.

\subsection{Domain Transfer in Genre Classification} 

In addition to establishing strong zero-shot ICL performance in genre classification, \newcite{kuzman2023automaticLLM} reported an out-of-distribution (OOD) gap when transferring trained BERT-sized models across datasets from different sources. Their findings were consistent with those reported for earlier PLM-based models, such as \newcite{lepekhin2022estimating}. 
In parallel work, \newcite{roussinov23emnlp} developed a specialized methodology to examine BERT-sized PLM OOD performance in genre classification using a topically diverse corpus and a topic model \cite{dieng20topic}. They also proposed a remedy based on synthetic augmentation, which reduced the OOD gap by a few percentage points on average. Additionally, \newcite{roussinov23emnlp} conducted what they described as a 'qualitative exploratory study' with the online interactive version of ChatGPT, suggesting that larger (GPT3-sized) models might also experience OOD performance gaps. However, they conducted a limited number of tests with LLMs and did not perform the tests of statistical significance. Our study here extends this work by methodologically confirming these gaps using two more recent and powerful families of LLMs (GPT-4.5 and Claude) accessed through their application interfaces, providing more controlled and replicable testing conditions compared to manual online interactions. Most importantly, we introduce a novel domain transfer approach by controlling the types of features used in classification, offering new insights into mitigating OOD performance gaps in LLMs.



\subsection{Generated Text Detection} 

The task of detecting text generated by large language models (LLMs) has become increasingly critical as models like GPT-3 and beyond produce more human-like content. Accurately distinguishing between human-written and AI-generated text is essential for curbing misinformation, maintaining academic integrity, and preserving content authenticity across platforms. Detection methods range from statistical analysis to watermarking and classifier-based systems, which can be fine-tuned or employed using zero-shot or few-shot In-Context Learning (ICL) approaches \cite{tang2024science, gehrmann2019gltr, kirchenbauer2023watermark}. The effectiveness of these methods often depends on the specific datasets and the presence of paraphrases. Paraphrasers, which subtly modify machine-generated content while preserving its meaning, further complicate detection efforts. While some works have identified the existence of an OOD gap in this task (e.g., \citealp{wang2024llm}), no universal solutions have been proposed.

\section{Conclusions and Further Work} 
\label{sec:org84e47fc}

Our key contribution lies in demonstrating that more careful prompt control In-Context Learning (ICL) can lead to enhanced performance in non-topical classification, particularly by enabling more effective domain transfer in genre classification and generated text detection. By introducing prompts to control which indicators should be prioritized or ignored, we achieved substantial improvements in ICL, reducing out-of-domain (OOD) performance gaps in LLMs by up to 20 percentage points across multiple topics. This method enables demonstration examples from one domain (e.g., \textit{sport}) to be successfully applied to another (e.g., \textit{science}), potentially reducing manual labeling costs and offering valuable insights for researchers and AI developers. 

Our innovation lies in the detailed control over classification criteria, systematically tested across two tasks. This approach yields notable improvements and, like CoT, its simplicity is a strength?offering a practical, effective method for enhancing LLM performance without extensive retraining or complex changes.

Our findings further highlight the superior performance of bigger and more advanced models like GPT-4o and Claude 3 Opus with detailed prompts in complex classification tasks. The ablation studies underscore the importance of each element in our method, confirming that prompt specificity plays a critical role and that the system remains robust even when prompts are paraphrased.

Looking ahead, more research is needed to explore cross-lingual capabilities. While previous research \cite{kuzman22ginco, ronnqvist21multilingual} has demonstrated that BERT-like models can be applied across languages, testing whether our approach to prompt control can extend to non-English texts remains a key challenge. This would require the development of a large multilingual corpus that spans a diverse range of genres and topics, opening up new opportunities for broader applicability.

\section{Limitations}
We considered two non-topical tasks, each evaluated with its respective dataset. To the best of our knowledge, no additional datasets are currently available for OOD exploration for these specific tasks. One key limitation of our study is the reliance on natural genre annotations, which may simplify the classification task, because natural genre labels can introduce superficial cues, such as formatting, that make classification easier than it would be with manually controlled labeling. This reliance raises questions about whether the model is learning the intended genre features or simply exploiting external characteristics. Additionally, our study is limited to English texts, and it remains unclear whether our findings would generalize to other languages with different linguistic structures and genre conventions. The lack of a suitable multilingual corpus with sufficient genre annotation and topical diversity restricts our current focus, though exploring non-English applications is a clear area for future work.

Another potential limitation is the sensitivity of our results to prompt variations. Although we designed prompts to ensure robustness, exhaustively testing all possible configurations is not feasible, and there may be subtleties in prompt phrasing that impact performance. While we observed that reducing topical criteria in outputs tends to improve accuracy, we cannot conclusively determine the underlying mechanics without further exploration.

We also recognize that our use of black-box LLMs from two commercial providers poses challenges to both generalizability and reproducibility. The black-box nature limits insight into the internal workings of the models, making it difficult to interpret how they process prompts and features. Additionally, these results can only be reproduced as long as the APIs for these LLMs remain available and stable over time. As an alternative, deploying LLMs on local clusters could provide more transparency and control, but this would require significant computational resources, which may not be readily available.

Finally, while our study focused on genre classification and generated text detection, future research could extend our approach to other non-topical classification tasks, such as sentiment analysis, author identification, or stylistic categorization. This would further validate and broaden the applicability of our findings.

\section{Ethical Impact}

The potential societal benefits of our findings are substantial, particularly in improving content moderation, information retrieval, and personalized recommendations. By enhancing the accuracy of genre classification and generated text detection, we can contribute to more efficient digital ecosystems, where content is categorized more effectively, misinformation is reduced, and educational tools are made more accurate. These improvements can positively impact user experiences, making digital platforms safer and more informative.

However, alongside these benefits come notable ethical risks. One significant concern is the possibility of reinforcing existing biases, especially if the training data lacks diversity or fails to represent a broad spectrum of perspectives. Such biases could lead to unfair outcomes, perpetuating stereotypes or marginalizing certain groups. As LLMs are increasingly used in various decision-making processes, the potential for such biased outputs to influence real-world outcomes becomes a critical issue that requires attention.

Another potential risk involves the control of indicators in prompts. While controlling which features LLMs prioritize can be a powerful tool for improving performance, it also opens the door to misuse. The same techniques that enhance genre classification and text detection could be exploited to bias outputs in other domains. For instance, in news generation or summarization, prompts could be manipulated to emphasize particular narratives or viewpoints, subtly shaping public opinion or spreading misinformation. The misuse of such controls could have far-reaching implications, especially in sensitive areas like media, politics, and public discourse.

To mitigate these risks, it is crucial to ensure transparency, fairness, and accountability in how indicator control is applied. Developing frameworks that guard against manipulation and bias, while promoting robustness and fairness, is essential to upholding the ethical use of generative AI technologies.

\bibliography{bibexport}

\appendix 
\section{{Appendix}}

\begin{table*}[t] 
\caption{Corpus of natural genre annotation \label{tabCorporaG}  (from \citealp{roussinov23emnlp}) } 
\centering
\small
\label{GenreTable}
\setlength{\tabcolsep}{2pt}
\begin{tabular}{llrl|l}
Genre & General prototypes & Texts & Natural source & Bias (\autoref{tabCorporaT})\\
\hline
ARGument & Expressing opinions, editorials & 126755 & Hyperpartisan \cite{kiesel19semeval} & Topics 9, 13\\
INSTRuction & Tutorials, FAQs, manuals & 127472 & A sample of StackExchange & Topics 19, 21\\
NEWS & Reporting newswires & 16389 & Giga News \cite{cieri02} & Topics 5, 9\\
PERSonal & Diary entries, travel blogs & 16432 & ICWSM set \cite{gordon09} &  Topic 23\\
INFOrmation & Encyclopedic articles & 97575 & A sample of Wikipedia &  Topics 1, 15, 20\\
Review & Product reviews & 1302495 & Amazon reviews \cite{blitzer07} & Topics 1, 16, 17\\
\hline
 & \textbf{Total} & 1687118 & \\ 
\end{tabular}
\end{table*}

\begin{table*}[h]
\caption{Keywords from ukWac for the topic model with 25 topics (from \citealp{roussinov23emnlp}) \label{tabCorporaT}
Seemingly similar names (e.g., Politics1 vs. Politics2) still have very different keywords (International vs Domestic). Additionally, any similarity does not pose a limitation to our method, as we split texts into on-topic/off-topic groups based on their topic-model scores (close vs distant from that topic), thus we are not relying on 'orthogonality' between the topics.
}
\centering
\small
\setlength{\tabcolsep}{2pt}
\begin{tabular}{lp{0.85\textwidth}} 
\textbf{Label}: \textbf{Nr} & \textbf{Top keywords} \\
Finances: 0 & insurance, property, pay, credit, home, money, card, order, payment, make, tax, cost, time, service, loan\\
Entertain: 1 & music, film, band, show, album, theatre, festival, play, live, sound, radio, song, dance, songs, tv, series\\
Geography: 2 & road, london, centre, transport, park, area, street, station, car, north, east, city, west, south, council, local\\
Business: 3 & business, management, company, service, customers, development, companies, team, experience, industry\\
University: 4 & students, university, research, learning, skills, education, training, teaching, study, work, programme\\
Markets: 5 & year, market, million, energy, waste, years, cent, industry, investment, government, financial, increase\\
Web: 6 & information, site, web, website, page, online, search, email, click, internet, details, links, free, find, sites\\
Science: 7 & data, research, system, analysis, model, results, number, time, science, methods, surface, cell, energy, test\\
*Cleaning: 8 & 2006, 2005, posted, 2004, june, july, october, march, april, september, 2003, august, january, november, post\\
Politics1: 9 & government, world, people, international, war, party, countries, political, european, country, labour, british\\
Travel: 10 & hotel, room, day, area, house, accommodation, holiday, visit, city, centre, facilities, town, great, tour\\
Health: 11 & health, patients, treatment, care, medical, hospital, clinical, disease, cancer, patient, nhs, risk, drug\\
Councils: 12 & development, local, community, council, project, services, public, national, planning, work, government\\
Life1: 13 & people, time, questions, work, make, important, question, problem, change, good, problems, understand\\
Software: 14 & software, system, file, computer, data, user, windows, digital, set, files, server, users, pc, video, mobile\\
Sports: 15 & game, club, team, games, play, race, players, time, season, back, football, win, world, poker, sports, sport\\
Religion: 16 & god, life, church, people, lord, world, man, jesus, christian, time, love, day, great, death, faith, men, christ\\
Arts: 17 & book, art, history, published, work, collection, world, library, author, london, museum, review, gallery\\
Law: 18 & law, act, legal, court, information, case, made, public, order, safety, section, rights, regulations, authority\\
Nature: 19 & food, water, species, fish, plants, garden, plant, animals, animal, birds, small, dogs, dog, tree, red, wildlife\\
History: 20 & years, century, house, st, john, royal, family, early, war, time, built, church, building, william, great, history\\
Engineering: 21 & range, design, light, front, high, car, made, water, power, colour, quality, designed, price, equipment, top\\
Politics2: 22 & members, meeting, mr, committee, conference, year, group, event, scottish, council, member, association\\
Life2: 23 & time, back, good, people, day, things, make, bit, thing, big, lot, can, long, night, feel, thought, great, find\\
School: 24 & people, children, school, support, young, work, schools, child, community, education, parents, local, care\\
\end{tabular}
\end{table*}

\begin{table*}[h]
\caption{Example of "Basic Prompt" prompt used in our study for genre classification. Class 1 is "Review". Class 2 is "PERSonal." Neither class labels nor descriptions are included in our prompts since we are looking at few-shot classification (solely from examples). The topic is "Entertainment". The examples are off-topic. The test text included is "Review" (Class 1). Punctuation and numbers have been restored in the texts for better readability.  Some parts of the prompt have been replaced with "..." for compactness. }
\small
\begin{tabular}{|p{.95\textwidth}|} 
\hline
\label{SimplestCoTPromptExample}

Classification Task

You are provided with example texts from two different classes. Your task is to classify a series of test texts into either Class 1 or Class 2 based on the characteristics observed in the example texts.

Here are some example texts of Class 1:

Example 1: This may be my second favorite Bill Bryson book, just behind "In a Sunburned Country." As I mentioned in his forum, it's beyond belief that he and his overweight friend, good old Katz, walked as far as they did in the wilderness and survived. It is a wonderful, relaxing, edifying, laugh-riot of a book as only Bill can provide. I love his work, and trust me, buy this?you won't be able to put it down. Even reading about the dangers will make you want to go hiking immediately...

Example 2: ...

...

Here are some example texts of Class 2:

Example 1: He eventually calmed down a bit, but it was clear he had hurt his arm somehow. He kept holding onto it with his other hand, keeping it very still and close to his body. The most telling sign that something was wrong was that he just sat there on the sofa. Hoonie never just sits there. He might sit somewhere and spit juice out of his mouth, or sit somewhere else while banging our fine Ikea furniture with his wooden hammer?but he never just sits. You know what I mean?...

Example 2: ...

...

Test Texts for Classification:

1: As a serious form of music outside of Jamaica, reggae stands on par with American and British rock and roll and R\&B. Before this breakthrough, reggae was often dismissed, despite Jimmy Cliff's "The Harder They Come" setting the stage. However, "Catch a Fire" clinched reggae's status. The album is a solid classic and a masterpiece, featuring lead vocals not only from Bob Marley but also from bandmate Peter Tosh on tracks like "400 Years" and "Stop That Train."...

...

Instructions:

First, based on the examples of texts of Class 1 and texts of Class 2 above, list at least three criteria by which Class 1 and Class 2 texts are different from each other. Next, apply those criteria to the test texts above to classify each of the test texts above into either Class 1 or Class 2. 

\end{tabular}
\end{table*}

\begin{table*}[h]
\caption{Example of "Prompt with simple Control" used in our study for genre classification. Same topic and genres as in the previous table. All the differences from the prompt in the previous table are marked with \textbf{bold}. }
\small
\begin{tabular}{|p{.95\textwidth}|} 
\hline
\label{CoTwithControlPromptExample}

Classification Task

You are provided with example texts from two different classes. Your task is to classify a series of test texts into either Class 1 or Class 2 based on the characteristics observed in the example texts.

Here are some example texts of Class 1:

Example 1: This may be my second favorite Bill Bryson book, just behind "In a Sunburned Country." As I mentioned in his forum, it's beyond belief that he and his overweight friend, good old Katz, walked as far as they did in the wilderness and survived. It is a wonderful, relaxing, edifying, laugh-riot of a book as only Bill can provide. I love his work, and trust me, buy this?you won't be able to put it down. Even reading about the dangers will make you want to go hiking immediately...

Example 2: ...

...

Here are some example texts of Class 2:

Example 1: He eventually calmed down a bit, but it was clear he had hurt his arm somehow. He kept holding onto it with his other hand, keeping it very still and close to his body. The most telling sign that something was wrong was that he just sat there on the sofa. Hoonie never just sits there. He might sit somewhere and spit juice out of his mouth, or sit somewhere else while banging our fine Ikea furniture with his wooden hammer?but he never just sits. You know what I mean?...

Example 2: ...

...

Test Texts for Classification:

1: As a serious form of music outside of Jamaica, reggae stands on par with American and British rock and roll and R\&B. Before this breakthrough, reggae was often dismissed, despite Jimmy Cliff's "The Harder They Come" setting the stage. However, "Catch a Fire" clinched reggae's status. The album is a solid classic and a masterpiece, featuring lead vocals not only from Bob Marley but also from bandmate Peter Tosh on tracks like "400 Years" and "Stop That Train."...

...

Instructions:

First, based on the examples of texts of Class 1 and texts of Class 2 above, list at least three criteria by which Class 1 and Class 2 texts are different from each other \textbf{in terms of genre (writing style), but not in topics or length.} Next, apply those criteria to the test texts above to classify each of the test texts above into either Class 1 or Class 2.  

\end{tabular}
\end{table*}

\begin{table*}[h]
\caption{Example of "Prompt with Detailed Control" used in our study for genre classification. Same topic and genres as in the previous table.  }
\small
\begin{tabular}{|p{.95\textwidth}|}
\hline
\label{DetailedCoTwithControlPromptExample}

Classification Task

You are provided with example texts from two different classes. Your task is to classify a series of test texts into either Class 1 or Class 2 based on the characteristics observed in the example texts. To accurately perform this task, please pay attention to the following aspects in each class:

{Formality and Tone: Notice whether the texts are formal or informal, professional or conversational.

Structure and Flow: Observe how the information is organized and presented. Is it narrative, following a linear progression, or is it segmented, perhaps formatted as questions and answers?

Complexity of Language: Assess the complexity of the language used, including the presence of specialized terminology or simpler, more general language.

Purpose and Interaction: Determine whether the text aims to inform, report, instruct, or facilitate a dialogue or interaction.

Sentence Structure: Look at the length and construction of sentences - are they typically complex with multiple clauses, or are they shorter and more directive?

Use of first- , third-, or second- person perspectives.

Use of active voice.

Soliciting feedback or further questions.

Dialogue-driven style.

Use of citations and references to studies.

Sharing personal experiences, or giving step-by-step advice.

Direct questions to the reader or community, and responses to hypothetical scenarios.

Use these principles to guide your classification, analyzing how each text aligns with the patterns observed in the example texts.

Important Note:

You SHOULD NOT be using topical content or size of the texts for classification!  The focus should be on how the texts are written, not what they are about. Do not mention specific fields or areas such as business, finances, entertainment, universities, markets, science, politics, travel, health, councils, software, sports, religion, arts, law, nature, history, engineering, school, etc. in your analysis. The classification should be universally applicable to any text based on the listed stylistic and structural elements.
Even when (almost) all the examples belong to the same topic, your criteria should not mention any specific topics and should be applicable to the texts on ANY topic!}

Here are some example texts of Class 1:

Example 1: ...

...

Instructions:

First, based on the examples of texts of Class 1 and texts of Class 2 above, list at least three criteria by which Class 1 and Class 2 texts are different from each other in terms of genre (writing style), but not in topics or length. Next, apply those criteria to the test texts above to classify each of the test texts above into either Class 1 or Class 2.  
\end{tabular}
\end{table*}

\begin{table*}[h]
\caption{Example of "Basic Prompt" used in our study for generated text detection.  }
\small
\begin{tabular}{|p{.95\textwidth}|}
\hline

Here are some example texts of Class 1:

...

Here are some example texts of Class 2:

...

Below, there are 10 texts. Classify each of them into either Class 1 or Class 2 based on the examples above. Present your response in the list format as in the example below. No explanations are needed. There should be nothing else in the output, just this list.

Example of output:...

 1: <First Test Text> 

...
 
 10: <Last Test Text>
 
\end{tabular}
\end{table*}

\begin{table*}[h]
\caption{Example of "Prompt with Simple Control" used in our study for generated text detection.  }
\small
\begin{tabular}{|p{.95\textwidth}|}
\hline

Here are some example texts of Class 1:

...

Here are some example texts of Class 2:

...

Below, there are 10 texts. Classify each of them into either Class 1 or Class 2 based on the examples above. \textbf{When classifying, don't use the topic of the text as a criteria.}  Present your response in the list format as in the example below. No explanations are needed. There should be nothing else in the output, just this list.

Example of output: ...

 1: <First Test Text> 

...
 
 10: <Last Test Text>
 
\end{tabular}
\end{table*}

\begin{table*}[h]
\caption{Example of "Prompt with Detailed Control" used in our study for generated text detection.  }
\small
\begin{tabular}{|p{.95\textwidth}|}
\hline

Here are some example texts of Class 1:

...

Here are some example texts of Class 2:

...

Below, there are 10 texts. Classify each of them into either Class 1 or Class 2 based on the examples above. When classifying, don't use the topic of the text as a criteria. \textbf{You SHOULD NOT be using topical content or size of the texts for classification!  The focus should be on how the texts are written, not what they are about. The examples above can be limited to particular topics. However, your classification should be universally applicable to any text regardless of the specific topical area such as business, finances, entertainment, universities, markets, science, politics, travel, health, councils, software, sports, religion, arts, law, nature, history, engineering, school, etc.} Present your response in the list format as in the example below. No explanations are needed. There should be nothing else in the output, just this list.

Example of output: ...

 1: <First Test Text> 

...
 
 10: <Last Test Text>
 
\end{tabular}
\end{table*}

\end{document}